\documentclass[11pt]{article}
\usepackage[T1]{fontenc}
\usepackage[utf8]{inputenc}
\usepackage{lmodern}
\usepackage[margin=1in]{geometry}
\usepackage{amsmath,amssymb}
\usepackage{graphicx}
\usepackage{booktabs}
\usepackage{tabularx,array}
\usepackage{float}
\usepackage{caption}
\usepackage{enumitem}
\usepackage[hidelinks]{hyperref}
\setlist{leftmargin=*}
\title{AI Scientists Are Only as Good as Their Evidence: A Stratified Ablation of Proprietary Data and Reasoning Skills in Drug-Asset Valuation}
\author{Yinan Wang\\Noah AI Research\\\texttt{yinan.wang@noah.bio}}
\date{}
\begin{document}
\maketitle
\begin{abstract}
AI Scientist agents are often evaluated as if their capability were mainly a function of model
quality, prompting, or reasoning scaffolds. In specialized scientific and commercial domains,
however, the limiting factor may be the \textbf{evidence substrate}: the private, curated, long-tail
record on which the agent is allowed to reason. We test this hypothesis in drug-asset valuation, a
knowledge-intensive task where correct go/watch/no-go calls require target biology, competitive
pipeline, clinical precedent and business-development (BD) intelligence. We report a controlled
three-arm ablation on a production valuation agent: \textbf{A} (plain LLM analyst, web search only),
\textbf{B} (+ public structured tools and a 14-dimension valuation playbook, verifier, objectivity policy
and adversarial red-team), and \textbf{C} (+ the proprietary \textbf{Noah AI} database of curated pipeline,
trial and deal intelligence). Across a 13-asset benchmark and 28--30 scored asset-seed cells per
condition, the non-proprietary tool/skills layer improves how evidence is used: B raises
tier-in-range accuracy from 0.80 to 0.89 and modestly lifts objectivity from 3.16 to 3.30, correcting
a plain-agent failure mode that rejects every validated-mechanism white-space opportunity. But
the non-proprietary tool/skills layer does not remove the factual ceiling: under generous capability-superset accounting, A and B
recover only 0.25 and 0.38 of the curated gold competitive record, while C recovers 0.96; on
the curated long-tail subset, C reaches 0.93 vs. 0.26/0.30. We therefore introduce
completeness-aware decision utility. Raw blind-panel decision quality is similar for A and B
(7.01 vs. 6.96), but
\textbf{informed decision-quality = decision-quality $\times$ gold-coverage} is \textbf{7.43 for C vs. 1.76/2.57 for
A/B}. A ceiling analysis makes the bottleneck explicit: even a perfect non-proprietary-data report
(DQ = 10) would be capped at 3.83 by B's 0.38 coverage on utility-evaluable cells, below the
observed C score. The result is not
that reasoning scaffolds are unimportant; rather, they improve calibration and discipline while the
proprietary evidence substrate determines the upper bound of what the AI Scientist can know and
therefore decide. The arXiv package contains only the paper source and figures;
proprietary-data-derived key-generation materials are not redistributed.

\end{abstract}
\section{Introduction}
Evaluating whether a drug asset --- a (\emph{target}, \emph{modality}, \emph{indication}) triple --- is worth
pursuing is a canonical high-stakes analytical task: bringing a drug to market costs on the order
of \$2.6B per approval \cite{ref1}, clinical success rates are low and phase-dependent \cite{ref2}, and aggregate
R\&D productivity has declined for decades \cite{ref3,ref4} --- so the quality of early go/no-go and
in-/out-licensing calls carries large expected value. The answer depends on target biology, the
competitive pipeline, clinical precedent, commercial size, regulatory feasibility and
business-development (BD) dynamics, and the ground-truth payoff is unknown for years. Machine
learning is increasingly applied across drug discovery and development \cite{ref5}; LLM agents build on
foundation-model, retrieval, reasoning and tool-use methods \cite{ref6,ref7,ref8,ref9,ref10} and are attractive here because
the task is evidence-synthesis at scale, but two distinct enhancements are routinely conflated when
their value is claimed:

\begin{enumerate}
\item \textbf{Skills and public structured tools}: explicit playbooks that encode scoring rubrics, hard
   gates, an objectivity charter and an adversarial review step, plus public APIs that structure
   evidence collection --- i.e. \emph{how} a competent analyst should reason and what artifacts to produce.
\item \textbf{Proprietary data}: a licensed, curated commercial-intelligence database --- here \textbf{Noah AI},
   which continuously monitors and integrates 50,000+ pharmaceutical-company websites, journal
   articles, academic conferences and news media into a structured corpus of programs, trials and
   deals far beyond what ad-hoc open-web search surfaces.
\end{enumerate}
Practitioners typically observe the \emph{combined} system and attribute its quality to "the model",
"the prompt", or "the data" without an ablation. This paper asks a sharper question: \emph{for an
AI Scientist agent doing knowledge-intensive valuation, is the bottleneck the reasoning procedure or
the evidence substrate it can access?}

\textbf{Contributions.}

\begin{enumerate}
\item A clean operationalization of a three-arm ablation (A/B/C) over an identical production
   valuation agent, with the proprietary-data layer starved by environment control rather than
   prompt suggestion (Section 2).
\item A 13-asset benchmark \emph{stratified by decision archetype} so that each stratum predicts whether the
   relevant failure should be factual incompleteness, calibration error, or both (Section 3).
\item A metric suite that separates \textbf{reasoning discipline} (tier calibration, objectivity and
   blind-panel verdict quality) from the \textbf{evidence substrate} (coverage of a proprietary
   competitive/deal record, long-tail recall and coverage-limited decision utility) (Section 4).
\item Empirical evidence that skills/public tools improve calibration, but proprietary data sets the
   factual and decision-utility ceiling: the non-proprietary stack still sees only 0.38 of the curated
   gold competitive record, while the full system sees 0.96 and dominates completeness-aware decision quality
   (Section 5).
\end{enumerate}
\section{System and Conditions}
\textbf{Base system.} The agent runs a \texttt{default\_14dim} playbook: it plans, collects evidence from
public APIs (ClinicalTrials.gov, PubMed, OpenTargets, OpenFDA) and the web, scores eight headline weighted
dimensions (competition, clinical evidence, scientific rationale, differentiation, unmet need,
commercial, development feasibility, BD) on a 0--10 scale with cited evidence, applies deterministic
guardrails (a crowding cap and a modality--target accessibility gate), runs a citation verifier, and
emits an evidence-cited scorecard. An objectivity charter (eight laws: verdict-first BLUF;
evidence-tiering of drivers; calibrated valuation anchors; disconfirming evidence must move the
score; exit questions; completeness; hallucination/grounding control; actionability) is enforced by
an adversarial red-team pass that may only \emph{lower} the recommendation.

\textbf{The three arms.} All arms use the same backbone model and equal time/scope budgets; the manipulated
variables are non-proprietary tooling/skills and proprietary-data access. The execution environment is \textbf{Claude Code
with Opus 4.8 MAX} for all arms, so model capability is held constant while tool/data access varies.

\begin{itemize}
\item \textbf{A} \emph{(plain)}: a competent generalist analyst with web search only, explicitly forbidden from
  reading any project skill, tool or datum; given no rubric. A fair baseline, not a crippled one.
\item \textbf{B} \emph{(+skills+public tools)}: the full playbook, public APIs, verifier, objectivity charter and red-team, with
  the proprietary data \emph{starved} (\texttt{LOCAL\_DATA\_DIR} pointed at an empty directory so the
  commercial-intelligence tools return nothing); evidence comes from web + public APIs.
\item \textbf{C} \emph{(+skills+public tools+data)}: identical to B but the proprietary Noah AI exports are
  present and queried first. The competitive pipeline is queried at \textbf{target level} (competition
  is a target$\times$modality property; an indication-scoped query over-filters and can return zero for
  sparse indications).
\end{itemize}
Each arm writes to an isolated run directory; conditions never share state.

\section{Benchmark Dataset}
\textbf{Stratification.} We select 13 single-target assets grouped into five archetypes, each chosen
because its \emph{known} structure predicts which layer should matter (Table 1). The strata deliberately
include both directions of failure: spaces that look attractive but are traps (S3), and spaces that
look unattractive but are genuine opportunities (S2). Bispecific/dual-target assets were excluded
because their competitor-truth denominator (the full primary-target landscape) is not comparable to
single-target assets.

\begin{table}[H]
\centering
\small
\caption{Five decision archetypes.}
\begin{tabularx}{\textwidth}{>{\raggedright\arraybackslash}p{0.10\textwidth} >{\centering\arraybackslash}p{0.06\textwidth} >{\raggedright\arraybackslash}X}
\toprule
Stratum & n & Archetype and the layer it should expose \\
\midrule
S1 & 4 & \textbf{Validated-but-crowded me-too} (validation-paradox trap). De-risking dimensions are high \emph{because} the class is validated; correct call is \emph{watch/no\_go}. Tests whether data captures the full crowding. \\
S2 & 2 & \textbf{Validated mechanism $\times$ white-space indication} (genuine opportunity). Should be \emph{elevated} with a quantified must-prove differentiation. Tests calibration. \\
S3 & 3 & \textbf{Uncrowded but target-unvalidated} (false-positive trap): empty \emph{because} no one validated it. Tests whether the agent avoids a false GO on a high uncrowded score. \\
S4 & 2 & \textbf{Deal-sensitive}: exact deal terms and buyer saturation drive the call. Tests the data layer's BD precision. \\
S5 & 2 & \textbf{Biological void}: an antibody against an intracellular target. Correct call \emph{no\_go} via the accessibility gate. \\
\bottomrule
\end{tabularx}
\end{table}
The final scored rollup contains \textbf{28--30 asset--seed cells per condition} (A=30; B/C=28) after
incomplete cells are excluded; S4 has the most missing B/C coverage and is reported separately.

\textbf{Gold answer keys and circularity.} For each asset, the gold answer key was drafted from the
same proprietary commercial-intelligence family available to C (Noah AI/Citeline/BMT exports) and
then manually audited and enriched by industry experts. The key includes the competitor set (name,
sponsor, phase, each tagged \emph{web-famous} or \emph{long-tail}), discontinued set, curated marquee
transactions, must-cite facts, trap conditions and known real-world outcome. This is intentionally
not an independent clinical-payoff ground truth. Instead, \emph{R\_gold} measures how much of a curated
proprietary diligence record each arm can recover. The design is therefore a direct test of the
paper's thesis -- access to the proprietary evidence substrate -- but it should not be read as a
database-independent estimate of all real-world competitors. We retain \emph{R\_union} for future
extensions where independently sourced web/expert additions exceed the current gold set. The Noah AI
corpus available to C contains 5,322 programs, 16,547 trials and 476 deals; each asset uses the
target-specific slice.

\section{Metrics: Definitions and Analysis Methods}
We group metrics into three analytical axes: factual coverage, methodological/objectivity discipline
and decision quality. Auto-checkable metrics are computed by a deterministic scorer
against the keys; subjective metrics are scored by a \textbf{hard-blind 3-judge panel}, following
LLM-as-a-judge practice \cite{ref11}. For each asset the
three condition reports are \emph{anonymized}: condition labels, run IDs/paths and evidence \textbf{provenance}
(source-database names, URLs) are scrubbed to \texttt{[ref]}, and the three are \textbf{deterministically shuffled}
into "Report 1/2/3" with a private map the judges never read. \textbf{N = 3} independent judges score each
report (8 objectivity codes 0--4 + decision-quality 0--10) given the asset identity and factual ground
truth (to detect omissions/over-claims) but \emph{not} the expected tier, scoring each report independently
before comparing; we take the \textbf{mean} (and majority recommendation) and record inter-judge spread
(\texttt{dq\_std}, typically 0--0.8). Judge metrics are retained at report level only when the required
artifact exists (A free-form report; B/C scorecard); truncated B/C runs without scorecards are
excluded for those arms. \textbf{Blinding scope (honest):} this hard-blinds \textbf{B vs C} (both scorecards;
only provenance differed, now scrubbed); \textbf{A} is a free-form report vs B/C scorecards --- a structural
difference that cannot be erased, so A is partly identifiable and its scores are read alongside the
auto-metrics rather than as a perfectly blinded style comparison.

\textbf{Recall vs. curated gold (R\_gold); Axis 1.}
\emph{Definition:} fraction of the curated gold competitor set that a
condition names. \emph{Method:} truth-side matching --- a gold program counts as recalled if any reported
program matches it by drug name or development code (case/punctuation-normalized, alias-aware);
matching is name/code based, not sponsor based, to avoid cross-crediting distinct same-sponsor
programs (an early sponsor-based matcher was found to inflate A and deflate C; see Section 7).
\emph{Interpretation:} this is a curated-gold coverage signal. The C $-$ A/B gap estimates the factual
record made reachable by the proprietary evidence source. Cells with no gold competitor denominator
have R\_gold undefined and are omitted from recall means, while still contributing to non-recall
metrics when judgeable.

\textbf{Recall vs. union (R\_union); Axis 1.}
\emph{Definition:} recall against the curated gold set plus independently discovered \texttt{web\_extra} programs.
\emph{Method:} identical matching over the augmented key. \emph{Note:} \texttt{web\_extra} is empty in this study, so
R\_union = R\_gold; the hook is retained for extension.

\textbf{Long-tail recall; Axis 1.}
\emph{Definition:} recall restricted to the curated long-tail subset (web-famous = false: preclinical,
regional/China, unregistered; initially heuristic and expert-audited). \emph{Method:} as R\_gold on the
long-tail subset. \emph{Why:} the sharpest test
of the data layer --- these are hard to surface reliably via ad-hoc web search.

\textbf{Deal recall; Axis 1.}
\emph{Definition:} fraction of the gold deal set a condition cites. Deal gold keys are curated with
marquee real transactions (e.g. Merck/Prometheus \$10.8B, J\&J/Momenta \$6.5B) rather than
auto-extracted, which produced tangential rows. \emph{Method:} company/asset token matching between
cited and gold deals. \emph{Why:} BD attractiveness depends on knowing who transacted on what terms.

\textbf{Over-generation ("extra programs"); Axis 1.}
\emph{Definition:} count of reported programs not matched to the (non-exhaustive) gold set. \emph{Method:}
unmatched-found count. \emph{Caveat:} this is \textbf{not} a hallucination rate --- because the gold set is
non-exhaustive, a higher number (especially for C) mostly means \emph{more real programs surfaced}; true
grounding is the judge HALLU code.

\textbf{Objectivity; Axis 2.}
\emph{Definition:} adherence to the eight objectivity laws (verdict-first, evidence-tiering,
calibration, disconfirming-evidence-moves-score, exit questions, completeness, grounding and
actionability), scored 0--4 per law and averaged. \emph{Method:} blind LLM judge against a fixed rubric;
computed only where the required report artifact exists.

\textbf{Key-fact coverage; Axis 2.}
\emph{Definition:} fraction of the asset's must-cite facts present in the report. \emph{Method:} salient-token
coverage over the report text and stated claims.

\textbf{Tier-in-range; Axis 3.}
\emph{Definition:} the recommendation lies in the gold tier range (an honest \emph{requires\_review} on a
borderline asset counts as in-range). \emph{Method:} membership test against the key's tier range.

\textbf{False-GO rate; Axis 3.}
\emph{Definition:} fraction of assets recommended \emph{go} that should not be --- the cardinal error.
\emph{Method:} indicator over assets whose gold range excludes \emph{go}.

\textbf{Decision quality; Axis 3.}
\emph{Definition:} a blind judge's 0--10 rating of whether a skeptical chief-scientific-officer could act
on the report (specific, decision-useful, names the next experiment/catalyst). \emph{Method:} blind LLM
judge; report-artifact-present runs only.

\textbf{Informed decision-quality; Axis 3.}
\emph{Definition:} decision-quality $\times$ R\_gold. \emph{Motivation:} raw DQ measures verdict-soundness on the
report prose alone; a blind judge cannot see missing competitors, so DQ is insensitive to
completeness (empirically, A $\approx$ B on DQ despite 4$\times$ recall gap). A decision taken on \textasciitilde{}25\% of the
competitive field is lower-quality even if the verdict is sound, so informed-DQ folds coverage into
the quality metric. Operationally, informed-DQ is averaged only over cells where both blind-panel DQ
and R\_gold are defined; raw DQ/objectivity keep their own judgeable-report denominator.

\textbf{Coverage-limited decision utility; Axis 3.}
\emph{Definition:} the maximum decision utility allowed by the evidence substrate. We report
informed-DQ = DQ $\times$ R\_gold as the main proxy, and we also report three robustness views: a hard
ceiling if DQ were perfect (10 $\times$ R\_gold), a pessimistic min proxy (10 $\times$ min(DQ/10, R\_gold)), and a
geometric proxy (10 $\times$ sqrt((DQ/10) $\times$ R\_gold)). These are not claims that product utility is the only
correct utility function; they test whether C's advantage survives reasonable ways of saying that a
good scientific decision requires both sound reasoning and adequate evidence.

\textbf{Cost telemetry.} We collect wall-clock, output-token and tool-call telemetry when available, but
the present analysis does not make cost a load-bearing claim because per-condition accounting is
incomplete in this run. Qualitatively, A is a single web analyst, whereas C can replace many web
searches with offline local reads; this should be measured rigorously in follow-up work.

\textbf{Capability-superset accounting for factual capacity.} Each arm's tool-set strictly contains the
previous: A (web) $\subset$ B (web + public APIs + skills) $\subset$ C (+ proprietary data). For factual found-sets
(competitors and deals), the main table asks a \emph{capability} question: what could each stack recover
if a higher arm retained all lower-arm factual discoveries? We therefore credit each arm with the
cumulative union of its own and all lower arms' findings, making factual recall monotone in
capability (A $\leq$ B $\leq$ C). This is deliberately generous to B and C and should be read as a capacity
estimate, not a claim that every autonomous C run actually re-finds every A web citation. The
interpretation most relevant to this paper is conservative: even under this generous accounting, the
non-proprietary stack reaches only 0.38 coverage, while the data-enabled stack reaches 0.96. Judge,
objectivity and decision-quality metrics are scored on each report's own prose and are never unioned.

\textbf{Protocol.} Same model and budget across arms; frozen data snapshot; isolated run
directories; blind judging without the expected tier; 2--3 seeds per asset; results reported as
bands, not decimals (run-to-run noise on a dimension is \textasciitilde{}$\pm$0.3).

\section{Results}
\subsection{Overall}
\begin{table}[H]
\centering
\scriptsize
\caption{Overall A/B/C comparison (A n=30, B/C n=28 asset--seed cells; judge metrics averaged over reports that passed artifact checks; paired sensitivity in Table 4). Factual recall uses capability-superset accounting; judge metrics do not. Undefined factual denominators are omitted from recall/informed-DQ means. Bold marks the best arm per metric.}
\begin{tabularx}{\textwidth}{>{\raggedright\arraybackslash}X >{\centering\arraybackslash}p{0.145\textwidth} >{\centering\arraybackslash}p{0.145\textwidth} >{\centering\arraybackslash}p{0.145\textwidth} >{\centering\arraybackslash}p{0.145\textwidth}}
\toprule
Metric & A plain & B +skills+public tools & C +skills+public tools+data & A$\rightarrow$C \\
\midrule
\multicolumn{5}{l}{\emph{Axis 1 --- factual grounding}} \\
Recall vs. \textbf{curated gold set} (R\_gold) & 0.25 & 0.38 & \textbf{0.96} & +0.71 \\
long-tail (not web-famous) & 0.26 & 0.30 & \textbf{0.93} & +0.67 \\
Phase-3 programs & 0.34 & 0.47 & \textbf{0.99} & +0.65 \\
Recall vs. union (= curated gold today) & 0.25 & 0.38 & \textbf{0.96} & +0.71 \\
Deal recall & 0.81 & 0.88 & \textbf{0.92} & +0.11 \\
Extra programs (not hallucination) & 5.5 & 15.6 & 22.9 & --- \\
\multicolumn{5}{l}{\emph{Axis 2 --- methodology / objectivity}} \\
Objectivity (/4, hard-blind panel) & 3.16 & 3.30 & \textbf{3.60} & +0.45 \\
Key-fact coverage & \textbf{0.93} & 0.86 & \textbf{0.93} & 0.0 \\
\multicolumn{5}{l}{\emph{Axis 3 --- decision quality}} \\
Tier-in-range & 0.80 & \textbf{0.89} & \textbf{0.89} & +0.09 \\
Decision quality (/10, panel) --- \emph{verdict-soundness} & 7.01 & 6.96 & \textbf{7.65} & +0.64 \\
\textbf{INFORMED decision-quality} (DQ $\times$ gold-coverage) & 1.76 & 2.57 & \textbf{7.43} & +5.67 \\
False-GO rate & 0.0 & 0.0 & 0.0 & 0.0 \\
\bottomrule
\end{tabularx}
\end{table}
\begin{figure}[H]
\centering
\includegraphics[width=0.95\textwidth]{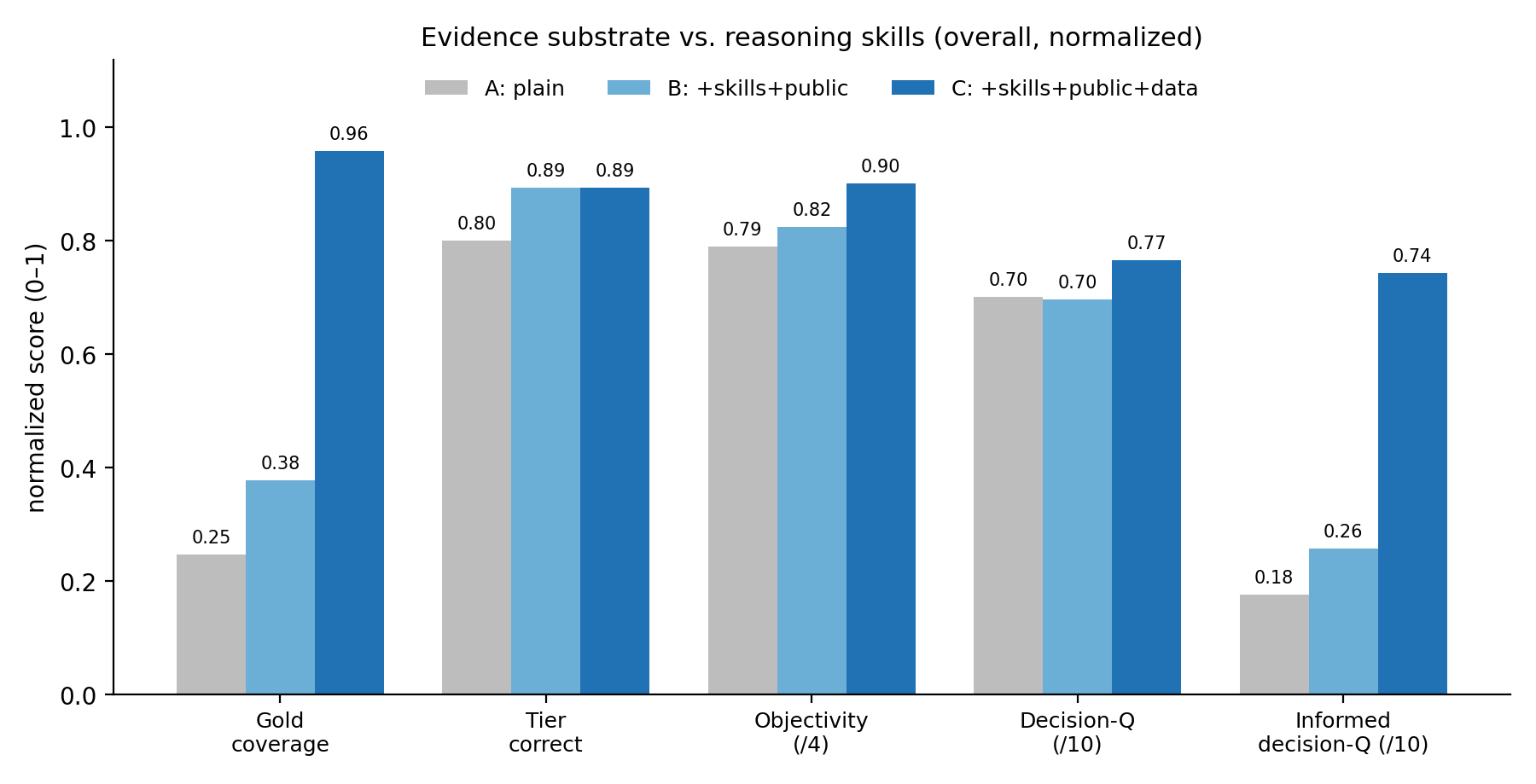}
\caption{A/B/C across the headline evidence and decision metrics (overall, normalized to 0--1: objectivity/4, decision-quality/10). B's skills/public tools lift tier-correctness and modestly lift objectivity; data (C) dominates coverage and informed decision-quality; B's verdict-soundness is similar to A (informed decision-Q B $\approx$ A).}
\end{figure}
\begin{table}[H]
\centering
\scriptsize
\caption{Completeness-limited decision utility. All rows use the same per-cell denominator: cells where both blind-panel DQ and R\_gold are defined (A n=22, B/C n=21). The key test is whether C remains superior if one changes the way coverage and verdict quality are combined.}
\begin{tabularx}{\textwidth}{>{\raggedright\arraybackslash}X >{\centering\arraybackslash}p{0.145\textwidth} >{\centering\arraybackslash}p{0.145\textwidth} >{\centering\arraybackslash}p{0.145\textwidth} >{\centering\arraybackslash}p{0.145\textwidth}}
\toprule
Utility view & A plain & B +skills+public tools & C +skills+public tools+data & Reading \\
\midrule
R\_gold coverage & 0.262 & 0.383 & \textbf{0.956} & Curated diligence record visible to the agent \\
Raw blind-panel DQ (/10) & 6.879 & 7.159 & \textbf{7.794} & Verdict soundness from report prose \\
Product proxy: DQ $\times$ R\_gold & 1.757 & 2.569 & \textbf{7.426} & Main informed-DQ result \\
Coverage ceiling if DQ = 10 & 2.615 & 3.831 & \textbf{9.563} & Per-cell maximum possible informed-DQ under each evidence substrate \\
Pessimistic min proxy & 2.539 & 3.450 & \textbf{7.500} & Per-cell decision limited by the weaker of DQ and coverage \\
Geometric proxy & 3.571 & 4.358 & \textbf{8.555} & Per-cell smooth penalty for either weak reasoning or weak coverage \\
\bottomrule
\end{tabularx}
\end{table}
This table is the central "evidence substrate" result. B improves the reasoning process, but at
0.383 coverage on utility-evaluable cells its best possible product-style informed-DQ is 3.83 even
with a perfect report. C's observed score is 7.43 because the proprietary data changes the factual
ceiling, not merely the wording of the recommendation.

\begin{table}[H]
\centering
\small
\caption{Retrieval-accounting sensitivity on fully paired cells. This table restricts to the 27 asset-seed cells where all three conditions produced scored outputs. "Raw" scores each condition's own reported found-set; "capacity" applies the capability-superset accounting used in Table 2. As above, recall rows omit cells with no gold competitor denominator. The data-enabled arm remains far ahead under both views.}
\begin{tabularx}{\textwidth}{>{\raggedright\arraybackslash}X >{\centering\arraybackslash}p{0.18\textwidth} >{\centering\arraybackslash}p{0.18\textwidth} >{\centering\arraybackslash}p{0.18\textwidth}}
\toprule
Metric on 27 paired cells & A plain & B +skills+public tools & C +skills+public tools+data \\
\midrule
Raw autonomous R\_gold & 0.262 & 0.233 & \textbf{0.929} \\
Capacity R\_gold & 0.262 & 0.383 & \textbf{0.956} \\
Raw long-tail recall & 0.271 & 0.078 & \textbf{0.905} \\
Capacity long-tail recall & 0.271 & 0.302 & \textbf{0.929} \\
Tier-in-range & 0.778 & \textbf{0.889} & \textbf{0.889} \\
\bottomrule
\end{tabularx}
\end{table}
\subsection{Per-dimension analysis (definition $\rightarrow$ method $\rightarrow$ conclusion)}
We restate each dimension's conclusion in light of the per-stratum decomposition (Table 5).

\begin{itemize}
\item \textbf{R\_gold (curated-gold coverage).} With a target-scoped query C recalls \textbf{0.96 overall} (reaching 1.0
  on S2--S4 and 0.97 on S1; the S5 shortfall of 0.78 reflects the intracellular-target strata where
  antibody-modality competitors are inherently sparser). The substantive quantity is the \textbf{A/B
  floor}: web-only (A) and non-proprietary tools/skills (B) surface only \textbf{0.25 / 0.38} of
  the curated competitive record. \textbf{Conclusion: 60--75\% of the curated competitive record is
  unreachable without the data (A$\rightarrow$C +0.71)}. B improves over A (0.38 vs 0.25, +0.13) via
  public-API structured-field extraction (ClinicalTrials.gov \texttt{Intervention(s)} parsing), but the
  marginal contribution of non-proprietary public tooling to coverage is modest compared to the data's +0.58 (B$\rightarrow$C).
\item \textbf{Long-tail recall.} C = 0.93 vs. B = 0.30, A = 0.26. \textbf{Conclusion: curated long-tail programs
  (preclinical/regional/unregistered) are not reliably surfaced without the proprietary feed};
  this is the cleanest single signal of data value.
\item \textbf{Deal recall.} Monotone under capability-superset capacity accounting: A 0.81, B 0.88, C 0.92.
  \textbf{Conclusion: each capability layer adds incrementally.} Web-famous marquee deals (large
  acquisitions) are highly web-visible, so even A recovers 81\%; B's public-API tools
  add 7pp, and the proprietary deal feed adds 4pp more. Deal recall is the most web-accessible
  factual axis --- and now monotone, correcting an earlier anomaly where C under-searched the web for
  historical acquisitions that A found. Because this row is unioned, it should be read as factual
  capacity rather than single-run retrieval behavior.
\item \textbf{Objectivity.} (hard-blind panel, eight-code rubric) C = 3.60, B = 3.30 vs. A = 3.16
  (+0.45 A$\rightarrow$C; +0.14 A$\rightarrow$B). \textbf{Conclusion: B's skills layer (verdict-first BLUF, evidence-tiering,
  exit questions, red-team) modestly improves audit discipline, but the effect is not uniform
  across strata; the data layer adds the larger gain (+0.31) because richer evidence enables more
  complete, grounded argumentation.}
\item \textbf{Tier-in-range.} B = C = 0.89 vs. A = 0.80. The entire gap originates in S2
  (Table 5, A = 0.0). \textbf{Conclusion: B partially fixes a systematic mis-calibration of the plain agent}
  (Section 5.3).
\item \textbf{Decision quality (verdict-soundness).} Panel DQ: A 7.01, B 6.96, C 7.65. Raw DQ is similar
  between A and B --- B's effect is concentrated in \emph{calibration} (tier) and \emph{audit discipline}
  (objectivity), not \emph{prose-level verdict-soundness}. C lifts DQ by +0.69 over B, the only arm
  that materially improves the quality rating.
\item \textbf{Does decision-quality reflect completeness? No --- by construction, and this matters.} Raw panel
  DQ scores \emph{verdict-soundness on the report prose}; a blind judge cannot see the missing long tail
  (it lives in the structured competitor list, not the summarizing scorecard). Completeness is the
  \textbf{recall axis} (C 0.96 vs A 0.25 / B 0.38). A decision taken blind to \textasciitilde{}75\% of the field is
  lower-quality even with a sound verdict, so we report \textbf{informed decision-quality = DQ $\times$
  gold-coverage: C 7.43 vs A 1.76 / B 2.57 (+5.67, \textasciitilde{}4$\times$)}. \textbf{Conclusion: data makes the
  decision \emph{informed} (C dominates \textasciitilde{}4$\times$); non-proprietary skills/tools make the verdict \emph{calibrated} (tier, objectivity);
  the non-proprietary stack remains coverage-limited.} Tables 3 and 4 show this is not an artifact
  of the product formula: C also leads under a hard ceiling, a min proxy and a geometric proxy.
\item \textbf{False-GO rate.} 0.0 for all arms. \textbf{Conclusion: no arm commits the cardinal error on this
  benchmark}; the discriminating decision error is \emph{under}-valuation (S2), not over-valuation. Even
  A is not fooled by the S5 biological-void cases (it knows an intracellular-target antibody is
  implausible), so the accessibility gate is a robustness backstop rather than the sole
  discriminator.
\item \textbf{Key-fact coverage.} A = 0.93, B = 0.86, C = 0.93. \textbf{Conclusion: key-fact coverage is
  comparably high across arms.} B's slight deficit likely reflects must-cite facts that are more
  readily surfaced by web search (A) or proprietary data (C) than by public-API sources alone.
\item \textbf{Over-generation.} C reports more unmatched programs (22.9) than B (15.6) or A (5.5).
  \textbf{Conclusion: not hallucination} --- against the non-exhaustive gold set these are predominantly
  \emph{real} programs beyond the key; higher numbers for C/B reflect more comprehensive evidence
  gathering. True grounding is judged by the blind HALLU code.
\end{itemize}
\begin{table}[H]
\centering
\scriptsize
\caption{Per-stratum decomposition. The tier gap is concentrated in S2; the gold-coverage gap is present everywhere. R\_gold/tier use auto-scored cells; DQ/objectivity use the judgeable-report subset and may have a smaller denominator, especially in S4. Where an asset has an empty gold competitor denominator, R\_gold is omitted for that cell (not scored as zero).}
\begin{tabularx}{\textwidth}{>{\raggedright\arraybackslash}p{0.30\textwidth} >{\centering\arraybackslash}p{0.15\textwidth} >{\centering\arraybackslash}p{0.14\textwidth} >{\centering\arraybackslash}p{0.14\textwidth} >{\centering\arraybackslash}p{0.14\textwidth}}
\toprule
Stratum (n) & R\_gold A/B/C & tier A/B/C & DQ A/B/C & obj. A/B/C \\
\midrule
S1 crowded (4, n=8) & 0.29/0.38/\textbf{0.97} & 1.0/1.0/1.0 & 7.5/7.3/\textbf{8.3} & 3.16/3.47/\textbf{3.68} \\
S2 white-space (2, n=6) & 0.38/0.63/\textbf{1.0} & \textbf{0.0}/0.5/0.5 & 4.9/5.8/\textbf{6.3} & 2.90/2.83/\textbf{3.33} \\
S3 unvalidated (3, n=9; R\_gold on 3 cells) & 0/0/\textbf{1.0} & 1.0/1.0/1.0 & 7.2/6.7/\textbf{7.7} & 3.19/3.29/\textbf{3.69} \\
S4 deal-sensitive (2 assets; cells 3/1/2) & 0.13/0.59/\textbf{1.0} & 1.0/1.0/1.0 & 6.7/7.0/\textbf{8.3} & 3.02/3.25/\textbf{3.63} \\
S5 biological void (2, n=4/4/3) & 0.23/0.23/\textbf{0.78} & 1.0/1.0/1.0 & 8.8/\textbf{9.0}/8.2 & 3.54/\textbf{3.78}/3.69 \\
\bottomrule
\end{tabularx}
\end{table}
\begin{figure}[H]
\centering
\includegraphics[width=0.95\textwidth]{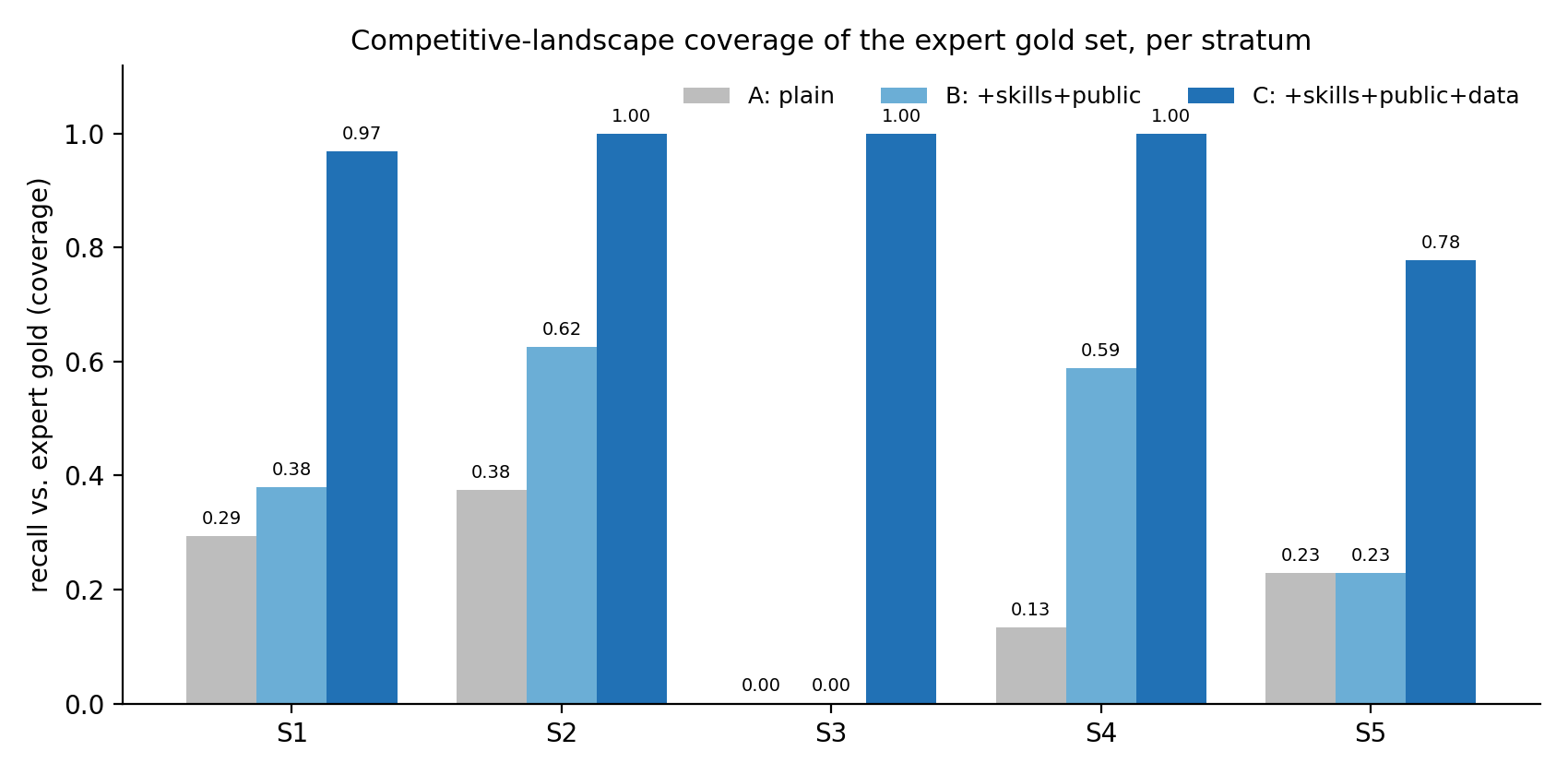}
\caption{Coverage of the curated gold competitive record per stratum. C reaches 0.96 overall (near the gold-coverage ceiling); web-only (A) and non-proprietary tools/skills (B) recover only a fraction --- the data layer's unique contribution.}
\end{figure}
\subsection{The calibration finding (S2)}
The plain agent's failure is \emph{indiscriminate pessimism}, not recklessness. It correctly rejects
crowded me-toos (S1) \emph{and} wrongly rejects validated-mechanism white-space plays (S2): across all
three seeds, A returns \emph{no\_go} on every S2 asset (IFNAR1 $\times$ scleroderma and IFNAR1 $\times$ lupus-nephritis;
tier-in-range = 0.0), unable to distinguish "avoid because crowded" from "pursue because the
mechanism is validated elsewhere and the indication is open." B partially recovers these (tier 0.5; the
scientific-rationale skill credits mechanism validation, and the calibrated thresholds elevate the
asset), improving panel decision quality over A (DQ 5.8 vs. 4.9; C reaches 6.3 with the fuller
record). This is the clearest evidence that the non-proprietary skills/tool layer changes the \emph{decision}, not merely the
presentation.

\section{Discussion}
The main result is an evidence-substrate bottleneck. The B layer changes how the agent reasons over the
facts it can see: they prevent the systematic under-valuation of S2 white-space opportunities and
improve audit discipline through verdict-first structure, evidence tiering and adversarial review. But they
do not make the missing competitive landscape visible. Even under capability-superset accounting
that is generous to the non-proprietary B arm, B recovers only 0.38 of the curated gold competitive record and
0.30 of curated long-tail programs. The proprietary data arm recovers 0.96 and 0.93 respectively.

This distinction matters for how AI Scientist systems should be evaluated. A report can be
methodologically sound and still be scientifically under-informed if it is written over the wrong
evidence base. In our runs, A and B have nearly identical raw blind-panel verdict quality
(7.01 vs. 6.96), yet they operate over a small fraction of the field. Table 3 makes the ceiling
explicit: with B's 0.383 coverage on utility-evaluable cells, even a perfect DQ=10 report would
have a product-style informed-DQ ceiling of 3.83, while the observed C system reaches 7.43. Under min and geometric
utility proxies, C remains the best arm. The conclusion is therefore not merely "data improves
recall"; it is that proprietary data changes the upper bound of decision quality.

This should not be read as an argument against reasoning scaffolds. Without the B layer, the agent is
miscalibrated on the validated-mechanism white-space stratum: it rejects every S2 opportunity. The
right architecture is layered: reasoning skills and public structured tools make the verdict calibrated and auditable; the
proprietary corpus makes the verdict informed. The weaker claim, "skills and data are both useful",
is true but undersells the result. For knowledge-intensive AI Scientist agents, non-proprietary
tooling and reasoning scaffolds are necessary control layers, while the evidence substrate is the
limiting resource.

Deal recall shows the boundary of the claim. Marquee transactions are unusually web-visible, so A
already recovers 0.81 and the data arm adds a smaller increment to 0.92 under capability-superset
capacity accounting. On famous assets and textbook biology the web baseline can be strong. The data
dependence appears precisely where one should expect it: long-tail pipeline programs, regional or
preclinical assets, sparse indications and diligence questions that require "who else exists" rather
than "what is the headline precedent".

\section{Threats to Validity}
\begin{itemize}
\item \textbf{No clean payoff ground truth.} Asset value settles over years; we proxy with auto-checkable
  facts, retrospective known outcomes and a blind judge, and anchor headline claims on the
  least-circular axes (data coverage, long-tail recall).
\item \textbf{Judge bias / blinding.} LLM judges can reward verbosity; we use a fixed adversarial rubric, a
  \textbf{hard-blind 3-judge panel} (provenance-scrubbed, shuffled, condition-labels stripped; mean of N=3;
  inter-judge std \textasciitilde{}0.5), independent-then-compare scoring, and supply factual ground truth. Blinding is
  full for \textbf{B vs C} but only partial for \textbf{A} (free-form vs scorecard structure cannot be erased).
  Judge metrics are not unioned or adjusted; small inversions (S5 DQ C $<$ A; S3 DQ B $<$ A) are within noise
  and reported honestly.
\item \textbf{Capability-superset accounting.} Factual recall in the main table is a capacity estimate that
  unions lower-arm discoveries into higher arms. This is generous to B/C and removes stochastic
  under-use of shared tools, but it is not the same as measuring single-run autonomous retrieval.
  The central data-dependence claim is robust to this generosity because B still reaches only 0.38
  coverage while C reaches 0.96.
\item \textbf{Completeness-aware utility.} Informed-DQ is a proxy, not a canonical utility theorem. We
  therefore report ceiling, min and geometric variants; all preserve the same qualitative ordering.
\item \textbf{Gold-set circularity.} The gold competitor/deal keys were seeded from the same proprietary
  commercial-intelligence family available to C and then expert-audited. This is appropriate for
  testing whether an agent can recover a curated proprietary diligence record, but it is not an
  independent clinical-payoff ground truth and favors the data-enabled arm on factual coverage.
  \texttt{web\_extra} is empty in this study, so R\_union currently equals R\_gold.
\item \textbf{Expert annotation bias.} The gold keys are expert-audited and curated rather than produced by
  an independent adjudication process, which improves domain validity but introduces curator
  judgment. Program-level recall claims should therefore be audited against the disclosed matching
  rules or re-scored against alternative keys when available.
\item \textbf{Small/uneven strata.} S4 has partial B/C scored-cell coverage (A/B/C = 3/1/2); S2/S3 at 3 seeds carry the
  load-bearing calibration claim, where the 0.0-vs-0.5 tier gap exceeds seed noise.
\end{itemize}
\section{Conclusion}
This study supports a simple thesis: for knowledge-intensive AI Scientist agents, capability is
bounded by the evidence substrate. Reasoning skills and public structured tools matter --- they help correct a systematic pessimism on
validated white-space opportunities and modestly lift objectivity from 3.16 to 3.30 --- but they cannot
manufacture the missing long-tail record. The non-proprietary tools/skills stack sees 0.38 of the curated gold
competitive record; the data-enabled agent sees 0.96. On the curated long-tail subset the gap is
0.30 vs. 0.93.

The consequence is visible at the decision level. Raw verdict-soundness is similar for A and B, but
a sound verdict over 25--38\% of the field is still poorly informed. On completeness-aware decision
quality, the full system reaches 7.43 vs. 1.76/2.57 for A/B, and even a hypothetical perfect
non-proprietary-data report is capped below C by its coverage deficit. The practical lesson is not "data
instead of skills"; it is "skills on top of the right data". For AI Scientist systems in domains
with proprietary, curated or long-tail evidence, the decisive capability may come less from another
reasoning prompt than from access to the corpus that defines what there is to reason about.

\textbf{Reproducibility and data availability.} The arXiv source package contains the paper source and
figures. The A (web-only) and B (public-API) arms are conceptually reproducible from public sources,
but the C arm and the gold-key drafting process depend on the \textbf{Noah AI} corpus and licensed
Citeline/BMT-derived exports, which are proprietary and cannot be redistributed. We therefore do
not include the key-generation script, proprietary-data-derived answer keys or per-asset run
artifacts in the arXiv source package. We report aggregate metrics, matching rules and the blind
judge rubric so the methodology can be audited and transferred to another curated
commercial-intelligence source.

\appendix
\section{Per-asset roster}
\begin{itemize}
\item \textbf{S1:} TL1A$\cdot$mAb$\cdot$UC, KLKB1$\cdot$mAb$\cdot$HAE, BAFF$\cdot$mAb$\cdot$SLE, FcRn$\cdot$mAb$\cdot$MG.
\item \textbf{S2:} IFNAR1$\cdot$mAb$\cdot$scleroderma, IFNAR1$\cdot$mAb$\cdot$lupus-nephritis.
\item \textbf{S3:} CLCN1$\cdot$small-molecule$\cdot$MG, SELPLG$\cdot$mAb$\cdot$UC, LANCL2$\cdot$small-molecule$\cdot$UC.
\item \textbf{S4:} ITGA4$\cdot$mAb$\cdot$UC, IL23A$\cdot$mAb$\cdot$psoriasis.
\item \textbf{S5:} IRF5$\cdot$mAb$\cdot$RA, RIPK2$\cdot$mAb$\cdot$IBD.
\end{itemize}
\section{Notes}
Recommendation tiers: \emph{go} (composite $\geq$ 7.5), \emph{watch} (5.5--7.5), \emph{no\_go} ($<$ 5.5), \emph{requires\_review}
(band straddles a boundary). Phase encoding: 0 preclinical, 1/2/3 Phase I/II/III, 4 approved.
All runs use Claude Code with Opus 4.8 MAX. The Noah AI corpus comprises 5,322 pipeline programs,
16,547 trials, 476 BD deals and 993 catalysts across all targets.

\end{document}